# SD-QA: Spoken Dialectal Question Answering for the Real World

**Fahim Faisal, Sharlina Keshava, Md Mahfuz ibn Alam, Antonios Anastasopoulos**
Department of Computer Science, George Mason University
`{ffaisal,skeshav,malam21,antonis}@gmu.edu`

## Abstract

Question answering (QA) systems are now available through numerous commercial applications for a wide variety of domains, serving millions of users that interact with them via speech interfaces. However, current benchmarks in QA research do not account for the errors that speech recognition models might introduce, nor do they consider the language variations (dialects) of the users. To address this gap, we augment an existing QA dataset to construct a *multi-dialect, spoken* QA benchmark on five languages (Arabic, Bengali, English, Kiswahili, Korean) with more than 68k audio prompts in 24 dialects from 255 speakers. We provide baseline results showcasing the real-world performance of QA systems and analyze the effect of language variety and other sensitive speaker attributes on downstream performance. Last, we study the fairness of the ASR and QA models with respect to the underlying user populations.[1]

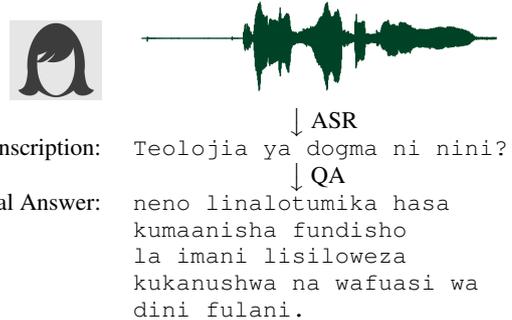

Figure 1: Illustration of the envisioned scenario for a user-facing QA system that SD-QA aims to evaluate (example from Kiswahili).

## 1 Introduction

The development of question answering (QA) systems that can answer human prompts with, or in some cases without, context is one of the great success stories of modern natural language processing (NLP) and a rapidly expanding research area. Usage of such systems has reached a global scale in that millions of users can conveniently query voice assistance like Google Assistant, Amazon Alexa, or Apple Siri through their smartphones or smart home devices. QA systems have also made large strides in technology-adjacent industries like healthcare, privacy, and e-commerce platforms.

Wide-spread adoption of these systems requires they perform consistently in real-world conditions, an area where Ravichander et al. (2021) note there is substantial room for improvement. Existing QA system evaluation benchmarks rely on text-based benchmark data that is provided in written format without error or noise. However, inputs to real-world QA systems are gathered from users through *error-prone interfaces* such as keyboards, speech recognition systems that convert verbal queries to text, and machine-translation systems. Evaluating production-ready QA systems on data that is not representative of real-world inputs is problematic and has consequences on the utility of such systems. For example, Ravichander et al. (2021) quantify and illustrate the effects of interface noise on English QA systems.

In this work, we address the need for realistic evaluations of QA systems by creating a multilingual and multi-dialect *spoken* QA evaluation benchmark. Our focus is the utility of QA systems on users with varying demographic traits like age, gender, and dialect spoken. Our contributions are as follows:

1. We augment the TyDi-QA dataset (Clark et al., 2020) with spoken utterances matching the questions. In particular, we collect utterances in four languages (Arabic, Bengali, English, Kiswahili) and from multiple varieties[2] (seven

---

[1]The dataset, model outputs, and code for reproducing all our experiments are available here: https://github.com/ffaisal93/SD-QA.

[2]We will use the terms "dialect" and "language variety"

for Arabic, eleven for English, and two for each of Bengali, Kiswahili, and Korean).

2. We perform contrastive evaluation for a baseline pipeline approach that first transcribes the utterances with an ASR system and then provides the transcription to the QA system. We compare general and localized ASR systems, finding wide divergences between the downstream QA performance for different language varieties.

## 2 The SD-QA dataset

The SD-QA dataset builds on the TyDi-QA dataset (Clark et al., 2020), using questions, contexts, and answers from five typologically diverse languages. Our focus is to augment the dataset along two additional dimensions. The first is a speech component, to match the real-world scenario of users querying virtual assistants for information. Second, we add a dimension on dialectal and geographical language variation.

### 2.1 Languages and Varieties

We focus on five of the languages present in the original TyDi QA dataset: English, Arabic, Bengali, and Kiswahili, and Korean. [3] In total, SD-QA includes more than 68k audio prompts in 24 dialects from 255 annotators. Table 1 presents a list of the different locations from where we collected spoken samples. A detailed breakdown of dataset and speaker statistics is available in Appendix A. Appendix B discusses the dialectal variation exhibited in all five languages.

We note that English and Arabic varieties are over-represented, but this is due to cost: it was easier and cheaper to source data in English and Arabic than in other languages; we plan to further expand the dataset in the future with more varieties for Bengali, Korean, or Kiswahili as well as more languages.

### 2.2 Data Collection Process

Data was collected through subcontractors. Each annotator was a native speaker of the language who grew up and lived in the same region we were focusing on. The annotators were paid a minimum of $15 per hour.[4] We aimed for gender-

| Language | Locations (Variety Code) |
|---|---|
| Arabic | Algeria (DZA), Bahrain (BHR), Egypt (EGY), Jordan (JOR), Morocco (MAR), Saudi Arabia (SAU), Tunisia (TUN) |
| Bengali | Bangladesh-Dhaka (BGD), India-Kolkata (IND) |
| English | Australia (AUS), India-South (IND-S), India-North (IND-N), Ireland (IRL), Kenya (KEN), New Zealand (NZL), Nigeria (NGA), Philippines (PHI), Scotland (SCO), South Africa (ZAF), US-Southeast (USA-SE) |
| Korean | South Korea-Seoul (KOR-C), South Korea-south (KOR-SE) |
| Kiswahili | Kenya (KEN), Tanzania (TZA) |

Table 1: Languages and sample collection locations (roughly corresponding to different spoken varieties) in the SD-QA dataset.

and age-balanced collection. The data for almost all dialects are gender-balanced,[5] but not all age groups are represented in all dialects (e.g. all Kenyan English and Jordan Arabic speakers are in the 18–30 age group).

For the collection of the Bengali and Kiswahili data we used the LIG-Aikuma mobile application (Gauthier et al., 2016) under the elicitation mode. The annotators were shown one question at a time, and they were instructed to first read the question in silence, and then read it out loud in a manner similar to how they would ask a friend or query a virtual assistant like Google Assistant, Siri, or Amazon Alexa.

**Data selection and Splits** We perform data selection and partitioning by following the process detailed in XOR-QA (Asai et al., 2020), another TyDi-QA derivative dataset. We use the development set of the original TyDi-QA as our test set, and randomly sample a part of the original TyDi-QA training set for our development set. The development and test partitions for the XOR-QA and our SD-QA dataset are exactly the same for Arabic and Bengali.[6] We follow the suggestions of Geva et al. (2019) and ensure that there is no

---
interchangeably.

[3]The languages were selected for their typological diversity and the wide range of variation they exhibit.

[4]We note, though, that no annotator needed more than 40 minutes for recording the maximum of 300 questions that corresponded to them.

[5]We note than none of our annotators self-reported as non-binary or other gender beyond male or female.

[6]English and Kiswahili are not part of XOR-QA.

annotator overlap between the development and the test set.

As our custom development set is constructed from the TyDi-QA training dataset, the SD-QA training set is constructed by discarding our development instances from the original TyDi-QA training data. We note, though, that we do not provide any spoken QA data for training.

### 2.3 Limitations

We believe that SD-QA certainly improves over existing benchmarks with regards to the accurate representation of the underlying users of QA systems in real-world scenarios. However, it is not without limitations. A user's location is a problematic proxy for their accent, and even lumping together users under a "variety" label can also be problematic, even if the users grew up in the same location. We asked the annotators to report additional metadata (e.g. other languages that they speak) that will facilitate future analysis to this end. The quality assurance process only focused on ensuring correct content, refraining from making judgements on annotators' accents.

Furthermore, since we lack access to *actual* spoken queries to smart assistants, we had to record readings of text questions. Read speech has different characteristics than spontaneous speech, especially in terms of rate and prosody (Batliner et al., 1995). In the future, we plan to investigate the creation of a spoken QA benchmark following a process similar to the Natural Questions dataset (Kwiatkowski et al., 2019) in order to produce an even more realistic benchmark.

## 3 Tasks and Evaluation

We perform three tasks over our dataset, defined below. The passage selection and minimal answer selection tasks are directly modeled after the primary task in TyDi-QA:

1. **Automatic Speech Recognition (ASR) Task:** A standard task defined over the utterances of the different language varieties. Given the audio file of the utterance, the model has to produce an accurate transcription. Since our questions are parallel across all language varieties, SD-QA can be used for contrastive evaluation of the robustness of ASR systems across different varieties.
2. **Passage Selection Task:** Given the question and a number of candidate passages, this task asks the model to return the index of the passage containing the answer, or null if no such passage exists.
3. **Minimal Answer Selection Task:** Given the question and a single passage, the task is to return the start and end byte indices of the minimal span that completely answer the question, or a YES/NO answer if appropriate, or NULL if such answer does not exist.

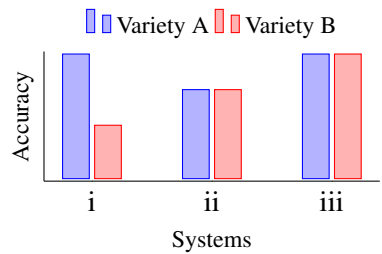

Figure 2: Schematic visualization of three systems tested on two language varieties. System (ii) is preferable to system (i) since it is more equitable. System (iii) is the ideal scenario.

**Evaluation** The two tasks that are similar to the TyDi-QA primary tasks use the same metric, $F_1$-SCORE.[7] For the ASR task, we evaluate the models computing the average word error rate (WER) across all test utterances. Unlike TyDi-QA, we (a) include English in our metrics, and (b) will not compute macro-averaged results across all languages as this measure could be biased towards English whose varieties are over-represented, but rather we only report macro-averages $\text{avg}_\mathcal{L}$ across the varieties of each single language. In addition, to measure the expected impact on actual systems users, we follow Debnath et al. (2021) in computing a population weighted macro-average ($\text{avg}_{\text{pop}}$) based on language community populations provided by Ethnologue (Eberhard et al., 2019).

For the ASR task, we evaluate the models computing the average word error rate (WER) across all test utterances. Unlike TyDi-QA, we (a) include English in our metrics, and (b) only report macro-averages across the varieties of each single language. We elect not to compute macro-averaged results across all languages as this measure could be biased towards English whose varieties are over-represented.

In addition, for all tasks, we will compare the models' robustness and equitability across the varieties. A truly robust, equitable system would perform equally well for all language varieties that it is meant to be used for. Since our dataset is parallel across varieties, it is ideal for *constrastive* evaluations with meaningful comparisons. Conceptually, consider the three systems depicted in Figure 2 tested on two language varieties A and B. Systems (i) and (ii) have the same macro-averaged accuracy, but system (ii) is more equitable than system (i): the quality of system (i) in language variety B is worse than in variety A. A more equitable situation is scenario (ii) were the two varieties have about the same performance, but this is achieved at the expense of performance in variety A. Scenario (iii) is an ideal case, where the system can properly model all language varieties and is also equitable. We discuss below how we measure the systems' unfairness across language varieties.

---

[7]We direct the reader to (Clark et al., 2020) for details.

**Comparing the Systems' Unfairness** When evaluating multilingual and multi-dialect systems, it is crucial that the evaluation takes into account principles of fairness, as outlined in economics and social choice theory (Choudhury and Deshpande, 2021). We follow the least difference principle proposed by Rawls (1999), whose egalitarian approach proposes to narrow the gap between unequal accuracies.

A simple proxy for unfairness is the standard deviation (or, even simpler, a $\max - \min$ performance) of the scores across languages. Beyond that, we will measure a system's *unfairness* with respect to the different subgroups using the adaptation of generalized entropy index described by Speicher et al. (2018), which considers equities within and between subgroups in evaluating the overall unfairness of an algorithm on a population. The generalized entropy index for a population of $n$ individuals receiving benefits $b_1, b_2, \ldots, b_n$ with mean benefit $\mu$ is

$$\mathcal{E}^\alpha(b_1, \ldots, b_n) = \frac{1}{n\alpha(\alpha - 1)} \sum_{i=1}^{n} \left[ \left(\frac{b_i}{\mu}\right)^\alpha - 1 \right].$$

Using $\alpha = 2$ following Speicher et al. (2018), the generalized entropy index corresponds to half the squared coefficient of variation.[8]

If the underlying population can be split into $|G|$ disjoint subgroups across some attribute (e.g. gender, age, or language variety) we can decompose the total unfairness into individual and group-level unfairness. Each subgroup $g \in G$ will correspond to $n_g$ individuals with corresponding benefit vector $\mathbf{b}^g = (b_1^g, b_2^g, \ldots, b_{n_g}^g)$ and mean benefit $\mu_g$. Then, total generalized entropy can be re-written as:

$$\mathcal{E}^\alpha(b_1, \ldots, b_n) = \sum_{g=1}^{|G|} \frac{n_g}{n} \left(\frac{\mu_g}{\mu}\right)^\alpha \mathcal{E}^\alpha(\mathbf{b}^g)$$
$$+ \sum_{g=1}^{|G|} \frac{n_g}{n\alpha(\alpha - 1)} \left[ \left(\frac{\mu_g}{\mu}\right)^\alpha - 1 \right]$$
$$= \mathcal{E}^\alpha_\omega(\mathbf{b}) + \mathcal{E}^\alpha_\beta(\mathbf{b}).$$

The first term $\mathcal{E}^\alpha_\omega(\mathbf{b})$ corresponds to the weighted unfairness score that is observed *within* each subgroup, while the second term $\mathcal{E}^\alpha_\beta(\mathbf{b})$ corresponds to the unfairness score *across* different subgroups. This formulation allows us to also study the trade-off between individual and group-level (un)fairness.

In this measure of unfairness, we define the benefit as being directly proportional to the system's accuracy. For the QA tasks, we make the assumption that the benefit that the user receives is directly proportional to the quality of the answer as measured by $F_1$-SCORE, so we will use $b = F_1$-SCORE for each question/answer pair. If the system produces a perfect answer ($F_1$-SCORE=1) then the user will receive the highest benefit of 1. If the system fails to produce the correct answer ($F_1$-SCORE=0) then the user receives no benefit ($b = 0$) from the interaction with the system. For the speech recognition task, we will simply use $b = 1 - \text{WER}$ as the corresponding benefit. We treat each interaction separately, assuming they are random draws from the distribution of all possible user interactions.

As we will show in the results section that follows, these unfairness scores are very useful for comparing two systems that can be applied over diverse populations tagged with sensitive features.

## 4 Baseline Results and Discussion

**Baseline Models** We benchmark the speech recognition systems using the ASR models through the Google speech-to-text API.[9] For all language varieties, we follow a *pipeline* approach,

---

[8]The coefficient of variation is simply the ratio of the standard deviation $\sigma$ to the mean $\mu$ of a distribution.

[9]https://cloud.google.com/speech-to-text

where we first transcribe the audio utterance and use the output as the input (question) to the QA model. We leave end-to-end multimodal approaches that operate directly on the audio for future work.

Our QA model follows the recipe of (Alberti et al., 2019), training a single model for all languages using multilingual BERT (Devlin et al., 2019) as a feature extractor. The model is trained for a maximum of 10 epochs, selecting the best checkpoint based on development set $F_1$-SCORE.

### 4.1 Speech Recognition Results

In this section, we discuss the quality of the outputs produced by different Automatic Speech Recognition (ASR) units. While this section only discusses transcription quality, Appendix D uses the framework from Section §2 to also evaluate the systems on their cross-dialect unfairness and across protected attributes (gender, age).

For English, we transcribe the regional variety utterances using both the localized[10] ASR model e.g. using the `en-AU` system for Australian English and the `en-NZ` for the data from New Zealand. In addition, we also transcribe all English data with the `en-US` system, which will allow us to compare the effectiveness of using localized models versus using a single "general" model. For Arabic we only use the localized model for each variety (as there is no "general" Arabic model available). For Kiswahili and Bengali we use localized models corresponding to the two data collection locations (Kenya-Tanzania and Bangladesh-India respectively) on both collections.

Table 2 presents the WER of the ASR models on the development set for all language varieties in SD-QA. The first important observation is that the average quality between languages varies significantly. The ASR systems achieve the lowest average WER on English (around 11-12), followed by Bengali (∼27.6) and Korean (24.4). Arabic and Kiswahili still prove challenging, with WER around 36 and 43.5 respectively.

Furthermore, we observe that the different dialects are not handled equally well by the models, even when we use localized models.[11]

For example, Indian English are consistently worse than other varieties (cf. WER over 13), while perhaps unsurprisingly the best-handled English variety is the one from the United States.

The comparison between the "general" `en-US` model and the different localized models reveals interesting divergences. When transcribing English from New Zealand, Nigeria, Scotland, South Africa, and Kenya, the localized models perform better than the US one. For Australian, Irish, and Philippine English, the differences between the two models are minor.

We also observe differences between the handling of Arabic varieties, with Algerian (DZA) proving particularly challenging. For both Bengali and Kiswahili, the two localized models produce exactly the same transcriptions. The quality is consistent for the two Bengali varieties around WER 27, with the Indian one being slightly better. On the other hand, the ASR quality of the two Kiswahili varieties exhibits the largest difference, with the Kenyan dataset receiving a 22% lower WER (cf. 37.9 and 49.1) than Tanzanian. Since the choice of the model does not affect downstream WER for Bengali and Kiswahili, we will only use one model in subsequent experiments.

### 4.2 QA Tasks Results

In this section, we investigate the effect of ASR errors on the downstream QA tasks, also discussing the performance across different language varieties. We first present results on the development set, on which we base our discussion, before providing results on the test sets.

**Passage Selection Task** Table 3 lists the obtained $F_1$-SCORE over the SD-QA development set, using both the original questions ("Gold") which simulate a perfect ASR system, as well as noisy transcriptions from different ASR models.

In almost all cases, using the noisy ASR output question has detrimental effect to the task. The reduction in $F_1$-SCORE varies from just 0.3 (in Indian Bengali) to 13 percentage points (in Tanzanian Kiswahili). The effect is generally less pronounced in Bengali (reduction of -0.8 points on average) and Arabic (-0.7 points), while it is significant in Kiswahili (-12.1 points on average).

Interestingly, there are three cases where the "noisy" ASR transcripts lead to slightly better performance that the gold transcripts. These are

---
[10]We will use the term "localized systems" as ones that are advertised as such to perform better on a specific language variety. We however do not have access to the actual training data to confirm this.

[11]We reiterate that because the dataset is *parallel* across the language varieties, these results are directly comparable.

| Model | English Variety | | | | | | | | | | | $\text{avg}_\mathcal{L}$ | $\text{avg}_{\text{pop}}$ |
| --- | --- | --- | --- | --- | --- | --- | --- | --- | --- | --- | --- | --- | --- |
| | AUS | IND-S | IND-N | IRE | NZL | NGA | PHI | SCO | ZAF | KEN | USA-SE | | |
| en-US | 10.94 | 14.25 | 13.96 | 9.43 | 14.62 | 12.67 | 12.67 | 11.21 | 11.55 | 13.68 | 8.97 | 12.17 | 12.29 |
| en-VAR | 10.41 | 17.27 | 17.34 | 9.77 | 8.69 | 11.10 | 12.28 | 9.30 | 8.73 | 11.03 | | 11.35 | 12.26 |

| Model | Bengali Variety | | | |
| --- | --- | --- | --- | --- |
| | BGD | IND | $\text{avg}_\mathcal{L}$ | $\text{avg}_{\text{pop}}$ |
| bn-BD | 28.40 | 26.73 | 27.56 | 27.85 |
| bn-IN | 28.47 | 26.73 | 27.6 | 27.90 |

| Model | Korean Variety | | | |
| --- | --- | --- | --- | --- |
| | KOR-C | KOR-SE | $\text{avg}_\mathcal{L}$ | $\text{avg}_{\text{pop}}$ |
| ko-KO | 27.95 | 24.63 | 26.29 | 27.76 |

| Model | Kiswahili Variety | | | |
| --- | --- | --- | --- | --- |
| | KEN | TZA | $\text{avg}_\mathcal{L}$ | $\text{avg}_{\text{pop}}$ |
| sw-KE | 32.6 | 42.31 | 37.46 | 37.69 |
| sw-TZ | 32.62 | 42.29 | 37.46 | 37.69 |

| Model | Arabic Variety | | | | | | | $\text{avg}_\mathcal{L}$ | $\text{avg}_{\text{pop}}$ |
| --- | --- | --- | --- | --- | --- | --- | --- | --- | --- |
| | DZA | BHR | EGY | JOR | MAR | SAU | TUN | | |
| ar-VARIETY | 38.41 | 36.29 | 35.87 | 36.05 | 35.82 | 36.50 | 35.61 | 36.36 | 36.42 |

Table 2: Development WER (lower is better) on each language variety, using different speech recognition models.

New Zealand English (NZL) using the localized ASR model and Algerian (DZA) and Morrocan (MAR) Arabic.

**Minimal Answer Task** Table 4 presents the $F_1$-SCORE on the minimal answer task. As before, in most cases using the noisy ASR transcriptions lead to QA performance deterioration.

In English, we find no difference between using the transcriptions of the general or the localized ASR systems (both have a macro-average $F_1$-SCORE of 33.3), which contrasts with our findings for the passage selection task where the better transcriptions from localized ASR system's lead to slightly better downstream performance.

Notably, the downstream performance for both Bengali dialects is improved when the input is the output of the ASR system, with a stronger effect for Indian Bengali. We plan on studying this interesting result in future work. On the other side of the spectrum, performance in Kiswahili is significantly impacted, with an average reduction of almost 14 percentage points.

**QA Systems Unfairness** We use the framework described in Section §2 to quantify the unfairness of the QA systems with respect to their underlying populations. The results are listed in Table 5. The overall unfairness across all 24 dialects is around 0.09 for both tasks. We observe the lowest unfairness score for Bengali speakers (∼0.04 in both tasks) and Korean, especially for the minimal answer task. English and Kiswahili speakers receive higher unfairness scores over 0.08, almost double the Bengali one. The higher scores for English (and Arabic) can be attributed to having several diverse dialects in our dataset that partition the population. In Swahili, this is due to the wide discrepancy between performance for Kenyan and Tanzanian speakers. This evaluation is useful for discerning which subgroups the models under perform in, highlighting that robustness and fairness improvements are necessary.

**The downstream effect of ASR noise** Perhaps unsurprisingly, we find that, *within each language*, the quality of the ASR transcription correlates with downstream QA performance, for both tasks. We calculate the Spearmans rank correlation coefficient for each language for both tasks, and visualize them in Figure 3 (also listed in Table 15 in Appendix F).[12] For English, we use the localized en-VAR models.

We note that Arabic vernaculars, unlike the other language varieties, seem to exhibit a different behavior, as evidenced by the lower correlation coefficients (c.f. -0.25 to $< -0.8$ for other languages in the minimal answer task). We leave an analysis of this behavior for future work, but we believe it can be attributed to the fact that the original questions are in Modern Standard Arabic, which may bias the speakers of different vernaculars to a varying extent.

In addition, we perform a linear mixed effect modeling (Bates et al., 2015) to find whether ASR transcription-downstream task correlations sustain while considering the effect of dialectal regions within a language. Here, we do not observe any significant transcription-downstream task correlation while using ASR word error rate (WER) as a fixed effect with WER conditioned on language/region as random effects. This can be attributed to the fact that inferior transcription quality adds noise that might affect the downstream prediction negatively, but, conversely, a perfect transcription does not guarantee correct predictions on the downstream task.

---
[12]The correlation coefficients are negative because *lower* WER is better, while *higher* $F_1$-SCORE is better.

| Transc. | AUS | IND-S | IND-N | IRE | NZL | NGA | PHI | ZAF | KEN | SCO | USA-SE | $\text{avg}_\mathcal{L}$ | $\text{avg}_{\text{pop}}$ | max-min |
|---|---|---|---|---|---|---|---|---|---|---|---|---|---|---|
| | | | | | **English Variety** | | | | | | | | | |
| Gold | | | | | | 53.4 | | | | | | | | |
| en-US | 51.1 | 50.9 | 50.2 | 51.9 | 52.4 | 49.9 | 50.3 | 51.6 | 48.7 | 51.5 | 52.8 | 51 | 50.7 | 4.5 |
| en-VAR | 52.8 | 48.8 | 48.8 | 52.2 | 54.2 | 51.8 | 50.9 | 52.9 | 51.3 | 52.6 | | 51.7 | 51.2 | 5.4 |

| Transc. | **Bengali Variety** | | $\text{avg}_\mathcal{L}$ | $\text{avg}_{\text{pop}}$ | max-min | Transc. | **Korean Variety** | | $\text{avg}_\mathcal{L}$ | $\text{avg}_{\text{pop}}$ | max-min |
|---|---|---|---|---|---|---|---|---|---|---|---|
| | BGD | IND | | | | | KOR-C | KOR-SE | | | |
| Gold | 57.9 | | | | | Gold | 54.9 | | | | |
| bn-XX | 56.6 | 57.6 | 57.1 | 57 | 1.0 | ko-KO | 48.0 | 48.4 | 48.2 | 48 | 0.4 |

| Transc. | **Kiswahili Variety** | | $\text{avg}_\mathcal{L}$ | $\text{avg}_{\text{pop}}$ | max-min |
|---|---|---|---|---|---|
| | KEN | TZA | | | |
| Gold | 69.2 | | | | |
| sw-XX | 57.5 | 56.3 | 56.9 | 56.9 | 1.2 |

| Transc. | DZA | BHR | EGY | JOR | MAR | SAU | TUN | $\text{avg}_\mathcal{L}$ | $\text{avg}_{\text{pop}}$ | max-min |
|---|---|---|---|---|---|---|---|---|---|---|
| | | | **Arabic Variety** | | | | | | | |
| Gold | | | | 65.0 | | | | | | |
| ar-VAR | 65.2 | 64 | 63 | 63.6 | 65.6 | 63.8 | 64.5 | 64.2 | 64.1 | 2.6 |

Table 3: Baseline passage selection results ($F_1$-SCORE, higher is better) on the SD-QA development set.

**Test Set Results** To facilitate future comparisons against both ASR and QA models, we also report test set results for the baseline pipeline approach. For English, we use the localized ASR models, as the dev set analysis shows they are better than the "general" US English model. Table 9 in Appendix C shows the ASR system's quality on the test set, Table 10 in Appendix C presents the results for all dialects on the passage selection task, while minimal answer task results are listed in Table 11 in Appendix C.

As in the development set, the noisy transcriptions lead to worse downstream performance compared to using the gold questions. Unlike the development set results though, we note that this hold for *all* languages and dialects for both tasks, even for e.g. Bengali (where the noisy transcriptions lead to slightly higher $F_1$-SCORE for the minimal answer task). In addition, the $F_1$-SCORE differences between the gold and noisy settings are generally larger than those we observed on the development set.

We refrain from performing any additional analysis on the test set, and suggest that any future analysis be conducted on the development set, so that all test set results reflect an estimation of real-world performance on *unseen* data.

## 5 Related Work

Due to space limitations, we provide a detailed account of related work on QA benchmark datasets and on multilingual QA data and approaches in Appendix H. We discuss here, though, the most relevant works on Spoken QA.

**Speech QA** A number of recent studies are done to bridge the gap between text based QA system and speech data. In Spoken SQuAD (Lee et al., 2018b), the authors propose a new task where the question is in textual form but the related reading comprehension is given in speech form. So transcription is performed on the speech data and the output can be in either text form or audio time span. Even using state-of-the-art speech transcription model, the authors show severe deterioration in performance. The authors further propose subword unit sequence embedding based mitigation strategies. This work was further extended to the ODSQA dataset (Lee et al., 2018a), where the question is also given in speech form. Another interesting study is DDNet (You et al., 2020), where the authors explore Spoken Conversational Question Answering (Spoken-CoQA). They used both speech and transcript in their feature vector embedding. To deal with the noise introduced by ASR units, they use knowledge distillation to minimize prediction loss computed using from speech based teacher model and transcript based student model.

| Transc. | AUS | IND-S | IND-N | IRE | English Variety NZL | NGA | PHI | ZAF | KEN | SCO | USA-SE | avg$_\mathcal{L}$ | avg$_{pop}$ | max-min |
|---|---|---|---|---|---|---|---|---|---|---|---|---|---|---|
| Gold | | | | | 37.3 | | | | | | | | | |
| en-US | 33.5 | 32.3 | 32.8 | 34.0 | 34.9 | 32.5 | 33.5 | 32.4 | 32.3 | 33.8 | 34.6 | 33.3 | 33.2 | 2.6 |
| en-VAR | 35.7 | 31 | 30 | 34.0 | 34 | 33.3 | 30.9 | 35.1 | 33.3 | 34.8 | | 33.3 | 32.7 | 4.8 |

| Transc. | Bengali Variety BGD | IND | avg$_\mathcal{L}$ | avg$_{pop}$ | max-min |
|---|---|---|---|---|---|
| Gold | 47.7 | | | | |
| bn-XX | 47.9 | 48.8 | 48.4 | 48.2 | 0.9 |

| Transc. | Korean Variety KOR-C | KOR-SE | avg$_\mathcal{L}$ | avg$_{pop}$ | max-min |
|---|---|---|---|---|---|
| Gold | 39.1 | | | | |
| kor-STD | 35.9 | 37 | 36.5 | 36 | 1.1 |

| Transc. | Kiswahili Variety KEN | TZA | avg$_\mathcal{L}$ | avg$_{pop}$ | max-min |
|---|---|---|---|---|---|
| Gold | 57.1 | | | | |
| sw-XX | 44.1 | 42.3 | 43.2 | 43.2 | 1.8 |

| Transc. | DZA | BHR | Arabic Variety EGY | JOR | MAR | SAU | TUN | avg$_\mathcal{L}$ | avg$_{pop}$ | max-min |
|---|---|---|---|---|---|---|---|---|---|---|
| Gold | | | 51.3 | | | | | | | |
| ara-VAR | 51.3 | 46.2 | 44.8 | 45.6 | 47.6 | 46.3 | 46.7 | 46.9 | 46.9 | 6.5 |

Table 4: Baseline results (F$_1$-SCORE) on the minimal answer task on the SD-QA development set.

| Language | Passage Selection Unfairness score ↓ | Avg. F$_1$-SCORE ↑ | Minimal Answer Unfairness score ↓ | Avg. F$_1$-SCORE ↑ |
|---|---|---|---|---|
| en-VAR | 0.082 | 51.7 | 0.078 | 33.3 |
| ar | 0.076 | 64.2 | 0.076 | 46.9 |
| bn | 0.043 | 57.1 | 0.047 | 48.4 |
| ko | 0.055 | 48.2 | 0.040 | 36.5 |
| sw | 0.089 | 56.9 | 0.090 | 43.2 |
| all | 0.089 | 55.9 | 0.093 | 39.6 |

Table 5: QA systems exhibit different levels of unfairness across languages, being more fair for Bengali speakers and less fair for Kiswahili or English speakers.

The most relevant work on the intersection of speech and QA is the recent NoiseQA (Ravichander et al., 2021) study, which explores the effect of noise in QA systems introduced by 3 main input interfaces: keyboard input, machine translation and speech input. Our task of assessing speech input error is conceptually similar to their speech input assessment. Their experiments show that the absence of punctuation in ASR outputs results in performance degradation by 5.1%. In addition, voice variation, accent and speaker's acoustic conditions as well as choice of ASR unit also play an important role. This study also shows that transcription of naturally spoken question results in errors like question type shift, ungrammatical or meaningless questions, corrupted named entities and dropped delimiters. Ravichander et al. (2021) make a number of recommendations including assessing the source of error, context-driven evaluation and community priorities while designing robust QA systems.

We believe that our work is a necessary expansion and complement of such studies. We go beyond Ravichander et al. (2021) by providing real audio data in more than one language instead of synthetic text-to-speech data. Furthermore, we go beyond English by providing data in four more languages and several varieties/locales.

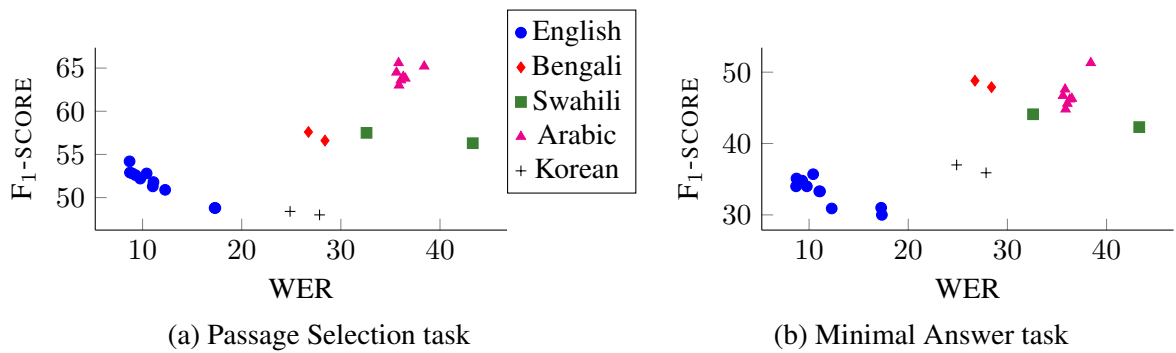

Figure 3: The downstream accuracy on the two QA task ($F_1$-SCORE) is generally negatively correlated to the question transcription quality (WER), for each language's dialects. This observation does not hold when comparing results across languages.

## 6 Conclusion

We present SD-QA, a new benchmark for the evaluation of QA systems in real-world settings, mimicking the scenario of a user querying a QA system through a speech interface. SD-QA is the largest spoken multilingual and multi-dialect QA dataset to date, with coverage of five languages and twenty-four dialects.

We provide baseline performance and fairness results on a pipeline that uses publicly available ASR models to transcribe spoken queries before passing them to a multilingual QA system. We showcase the QA systems' lack of robustness to noise in the question. We also discuss the *fairness* of both speech recognition and QA models with regards to underlying user characteristics, and show that a user's dialect can significantly impact the utility they may receive from the system.

Future areas of improvement for this work include expanding SD-QA to cover more languages and dialects, and additional analysis on attributes about the data and users. Ideally, we would like to discern which parameters most influence the performance of downstream language systems. We also plan to investigate the prospect of an end-to-end spoken QA system in an attempt to bridge the gap between the speech modality of the query and the textual modality of the currently available knowledge bases.

**Ethical considerations** All annotators involved in the collection of SD-QA signed informed consent forms about the nature of this work and were compensated fairly ($15/hour) for their work. The data collection process and its limitations are discussed in Section §2.2. The released dataset has been anonymized, while all demographic/identity characteristics are self-reported. The final release of the dataset includes an accompanying data statement (Bender and Friedman, 2018).

**Acknowledgments**

This work is generously supported by NSF Awards 2040926 and 2125466. The dataset creation was supported though a Google Award for Inclusion Research. We also want to thank Jacob Eisenstein, Manaal Faruqui, and Jon Clark for helpful discussions on question answering and data collection. The authors are grateful to Kathleen Siminyu for her help with collecting Kiswahili and Kenyan English speech samples, to Sylwia Tur and Moana Wilkinson from Appen for help with the rest of the data collection and quality assurance process, and to all the annotators who participated in the creation of SD-QA. We also thank Abdulrahman Alshammari for his help with analyzing and correcting the Arabic data.

## References

Abdel-Rahman Abu-Melhim. 1991. Code-switching and linguistic accommodation in arabic. In *Perspectives on Arabic Linguistics III: Papers from the Third Annual Symposium on Arabic Linguistics*, volume 80, pages 231–250. John Benjamins Publishing.

Chris Alberti, Kenton Lee, and Michael Collins. 2019. A BERT baseline for the natural questions.

Mikel Artetxe, Sebastian Ruder, and Dani Yogatama. 2019. On the cross-lingual transferability of monolingual representations.

Akari Asai, Jungo Kasai, Jonathan H Clark, Kenton Lee, Eunsol Choi, and Hannaneh Hajishirzi. 2020. XOR QA: Cross-lingual open-retrieval question answering. arXiv:2010.11856.

Douglas Bates, Martin Mächler, Ben Bolker, and Steve Walker. 2015. Fitting linear mixed-effects models using lme4. *Journal of Statistical Software*, 67(1):1–48.

Anton Batliner, Ralf Kompe, A Kießling, E Nöth, and H Niemann. 1995. Can you tell apart spontaneous and read speech if you just look at prosody? In *Speech Recognition and Coding*, pages 321–324. Springer.

Emily M. Bender and Batya Friedman. 2018. Data statements for natural language processing: Toward mitigating system bias and enabling better science. *Transactions of the Association for Computational Linguistics*, 6:587–604.

Mihaela Bornea, Lin Pan, Sara Rosenthal, Radu Florian, and Avirup Sil. 2020. Multilingual transfer learning for QA using translation as data augmentation. *CoRR*, abs/2012.05958.

Houda Bouamor, Nizar Habash, Mohammad Salameh, Wajdi Zaghouani, Owen Rambow, Dana Abdulrahim, Ossama Obeid, Salam Khalifa, Fadhl Eryani, Alexander Erdmann, et al. 2018. The MADAR Arabic dialect corpus and lexicon. In *Proceedings of the Eleventh International Conference on Language Resources and Evaluation (LREC 2018)*.

Eunsol Choi, He He, Mohit Iyyer, Mark Yatskar, Wen-tau Yih, Yejin Choi, Percy Liang, and Luke Zettlemoyer. 2018. QuAC: Question answering in context. *Proceedings of the 2018 Conference on Empirical Methods in Natural Language Processing*.

Monojit Choudhury and Amit Deshpande. 2021. How linguistically fair are multilingual pre-trained language models? In *Proc. AAAI*.

Jonathan H Clark, Eunsol Choi, Michael Collins, Dan Garrette, Tom Kwiatkowski, Vitaly Nikolaev, and Jennimaria Palomaki. 2020. TyDi QA: A benchmark for information-seeking question answering in typologically diverse languages. *Transactions of the Association for Computational Linguistics*, 8:454–470.

Arnab Debnath, Navid Rajabi, Fardina Fathmiul Alam, and Antonios Anastasopoulos. 2021. Towards more equitable question answering systems: How much more data do you need? In *Proceedings of the 59th Annual Meeting of the Association for Computational Linguistics and the 11th International Joint Conference on Natural Language Processing (ACL-IJCNLP)*. Association for Computational Linguistics.

Jacob Devlin, Ming-Wei Chang, Kenton Lee, and Kristina Toutanova. 2019. BERT: Pre-training of deep bidirectional transformers for language understanding. In *Proceedings of the 2019 Conference of the North American Chapter of the Association for Computational Linguistics: Human Language Technologies, Volume 1 (Long and Short Papers)*, pages 4171–4186.

David M Eberhard, Gary F Simons, and Charles D. (eds.) Fennig. 2019. Ethnologue: Languages of the world. 2019. online. Dallas, Texas: SIL International.

Elodie Gauthier, David Blachon, Laurent Besacier, Guy-Noel Kouarata, Martine Adda-Decker, Annie Rialland, Gilles Adda, and Grégoire Bachman. 2016. LIG-AIKUMA: a Mobile App to Collect Parallel Speech for Under-Resourced Language Studies.

Mor Geva, Yoav Goldberg, and Jonathan Berant. 2019. Are we modeling the task or the annotator? an investigation of annotator bias in natural language understanding datasets. In *Proceedings of the 2019 Conference on Empirical Methods in Natural Language Processing and the 9th International Joint Conference on Natural Language Processing (EMNLP-IJCNLP)*, pages 1161–1166, Hong Kong, China. Association for Computational Linguistics.

Nizar Y Habash. 2010. Introduction to arabic natural language processing. *Synthesis Lectures on Human Language Technologies*, 3(1):1–187.

Mandar Joshi, Eunsol Choi, Daniel Weld, and Luke Zettlemoyer. 2017. TriviaQA: A large scale distantly supervised challenge dataset for reading comprehension. *Proceedings of the 55th Annual Meeting of the Association for Computational Linguistics (Volume 1: Long Papers)*.

Mimi H Klaiman. 1987. Bengali. *The World's Major Languages, Croon Helm, London and Sydney*, pages 490–513.

Tom Kwiatkowski, Jennimaria Palomaki, Olivia Redfield, Michael Collins, Ankur Parikh, Chris Alberti, Danielle Epstein, Illia Polosukhin, Jacob Devlin, Kenton Lee, et al. 2019. Natural questions: A benchmark for question answering research. *Transactions of the Association for Computational Linguistics*, 7:453–466.

Chia-Hsuan Lee, Shang-Ming Wang, Huan-Cheng Chang, and Hung-Yi Lee. 2018a. ODSQA: Open-domain spoken question answering dataset. *2018 IEEE Spoken Language Technology Workshop (SLT)*.

Chia-Hsuan Lee, Szu-Lin Wu, Chi-Liang Liu, and Hung-yi Lee. 2018b. Spoken SQuAD: A study of mitigating the impact of speech recognition errors on listening comprehension. *Interspeech 2018*.

Patrick Lewis, Barlas Ouz, Ruty Rinott, Sebastian Riedel, and Holger Schwenk. 2019. MLQA: Evaluating cross-lingual extractive question answering.

Shayne Longpre, Yi Lu, and Joachim Daiber. 2020. MKQA: A linguistically diverse benchmark for multilingual open domain question answering.

Colin P Masica. 1993. *The indo-aryan languages*. Cambridge University Press.

Pranav Rajpurkar, Robin Jia, and Percy Liang. 2018. Know what you don't know: Unanswerable questions for SQuAD.

Pranav Rajpurkar, Jian Zhang, Konstantin Lopyrev, and Percy Liang. 2016. SQuAD: 100,000+ questions for machine comprehension of text.

Abhilasha Ravichander, Siddharth Dalmia, Maria Ryskina, Florian Metze, Eduard Hovy, and Alan W Black. 2021. NoiseQA: Challenge set evaluation for user-centric question answering.

John Rawls. 1999. *A Theory of Justice*. Harvard University Press.

Raj Reddy. 1989. Speech research at Carnegie mellon. In *Speech and Natural Language: Proceedings of a Workshop Held at Cape Cod, Massachusetts, October 15-18, 1989*.

Candide Simard, Sarah M Dopierala, and E Marie Thaut. 2020. Introducing the sylheti language and its speakers, and the soas sylheti project. *Language Documentation and Description*, 18:1–22.

Ho-Min Sohn. 2001. *The Korean language*. Cambridge University Press.

Till Speicher, Hoda Heidari, Nina Grgic-Hlaca, Krishna P Gummadi, Adish Singla, Adrian Weller, and Muhammad Bilal Zafar. 2018. A unified approach to quantifying algorithmic unfairness: Measuring individual & group unfairness via inequality indices. In *Proceedings of the 24th ACM SIGKDD International Conference on Knowledge Discovery & Data Mining*, pages 2239–2248.

Hanne-Ruth Thompson. 2020. *Bengali: A comprehensive grammar*. Routledge.

Jaehoon Yeon. 2012. Korean dialects: A general survey. *The languages of Japan and Korea*, pages 168–185.

Chenyu You, Nuo Chen, Fenglin Liu, Dongchao Yang, and Yuexian Zou. 2020. Towards data distillation for end-to-end spoken conversational question answering.

## A Dataset Details

We provide detailed statistics on the number of examples in the development and test sets, as well as on the speakers for each language/dialect in Table 6. The population statistics (Eberhard et al., 2019) for all dialectal regions are reported in Table 7.

## B Languages and Varieties

**Arabic** is practically a family of several varieties, forming a dialect continuum from Morocco to the west to Oman in the east. Most of NLP has focused on Modern Standard Arabic (MSA), the literary standard for language of culture, media, and education. However, no Arabic speaker is really a MSA native speaker–rather, most Arabic speakers in every day life use their native dialect, and often resort to code-switching between their variety and MSA in unscripted situations (Abu-Melhim, 1991). We follow a sub-regional classification for the Arabic dialects we work with, as these dialects can differ in terms of their phonology, morphology, orthography, syntax, and lexicon, even between cities in the same country. We direct the reader to (Habash, 2010) for further details on Arabic handling in NLP, and to (Bouamor et al., 2018) for a larger corpus of Arabic dialects.

The original TyDi-QA dataset provides data in MSA, as the underlying Wikipedia data are in MSA.[13] To capture the proper regional variation of Arabic as much as possible, the instructions to the annotators were to first read the question and then record the same question using their local variety, rephrasing as they deem appropriate.

**Bengali** is an Indo-Aryan language spoken in the general Bengal region, with the majority of speakers concentrated in modern-day Bangladesh and the West Bengal state in India (Klaiman, 1987). The language exhibits large variation, although languages like Sylheti and Chittagonian, once considered varieties of Bengali, are now considered as languages on their own (Simard et al., 2020; Masica, 1993). We focus our collection on two major geographical poles: Dhaka (Bangladesh) and Kolkata (India). The Dhaka variety is the most widely spoken one, while the Rarhi variety (Central Standard Bengali) is prevalent in Kolkata and has been the basis for standard Bengali. Thompson (2020) notes that the linguistic differences between the two sides of the border are minor. The differences are mainly phonological, although some lexical differences are also observed in common words. For example, Dhaka Bengali would use চইলা *(choila, to go)*, কইরা *(koira, to do)* instead of চলে *(chole, to go)*, করে *(kore, to do)* that Kolkata speakers would use. An Arabic influence is prevalent in Dhaka dialect, whereas the Kolkata dialect is more influenced by other Indo-Aryan languages, resulting in lexical differences (examples shown in Table 8).

**English** is a West Germanic Indo-European language with a very wide geographic distribution. This is attributed to colonialism, and resulted in large differences in grammatical patterns, vocabulary, and pronunciation between English varieties. For our corpus, it was only feasible to sample from a subsection of the English varieties that exist globally. We include regions where English is an "official" language (meaning it is generally the language of the government and of instruction in higher education): Australia, India, Ireland, Kenya, New Zealand, Philippines, Scotland, South Africa, and the United States. We note that there are important differences between English usage in these regions. For example, even though English is an official language in India and Kenya, speakers are more likely to use it as a second language, while having a different native language.

**Kiswahili** (or kiSwahili or Swahili, ISO code: swa) is a Bantu language that functions as a *lingua franca* for a large region of central Africa, as it is spoken in Tanzania, Kenya, Congo, and Uganda. Its grammar is characteristically Bantu, but it also has strong Arabic influences and uses a significant amount of Arabic loanwords. While there are more than a dozen Kiswahili varieties, the three most prominent are kiUnguja, spoken on Zanzibar and in the mainland areas of Tanzania, which is also the basis of considered-standard Kiswahili; kiMvita, spoken in Mombasa and other areas of Kenya; and kiAmu (or Kiamu), spoken on the island of Lamu and adjoining parts of the coast. Also prominent is Congolese Kiswahili (ISO code: swc), which we treat as a separate language because of its significant French influences (due to colonisation). In this work we collected utterances from Kenya (Nairobi area) and from Tanzania (Dar-es-Salaam area). Some of our contributors self-reported as non-native Kiswahili speakers, naming languages like Kikuyu (ISO code: kik) and Chidigo (ISO code: dig) as their native ones.

---
[13]While Clark et al. (2020) do not specify whether the data are in MSA or in any vernaculars, we rely on a native speaker for this observation.

| Language | Dialect | Development Set | | Test Set | |
|---|---|---|---|---|---|
| | | Examples | Speakers (M,F) | Examples | Speakers (M,F) |
| Arabic | Algeria (DZA) | 708 | 4 (2, 2) | 1380 | 7 (5, 2) |
| | Bahrain (BHR) | 708 | 3 (2, 1) | 1380 | 7 (5, 2) |
| | Egypt (EGY) | 708 | 3 (2, 1) | 1380 | 6 (4, 2) |
| | Jordan (JOR) | 708 | 3 (2, 2) | 1380 | 7 (4, 3) |
| | Morocco (MAR) | 708 | 4 (3, 1) | 1380 | 7 (4, 3) |
| | Saudi Arabia (SAU) | 708 | 3 (2, 1) | 1380 | 7 (5, 2) |
| | Tunisia (TUN) | 708 | 3 (2, 1) | 1380 | 8 (4, 4) |
| Bengali | Bangladesh (BGD) | 427 | 14 (8, 6) | 328 | 11 (5, 6) |
| | India (IND) | 427 | 4 (3, 1) | 326 | 2 (1, 1) |
| English | Australia (AUS) | 1000 | 4 (2, 2) | 1031 | 5 (1, 4) |
| | India-South (IND-S) | 1000 | 5 (4, 1) | 1031 | 4 (4, 0) |
| | India-North (IND-N) | 1000 | 4 (3, 1) | 1031 | 4 (–, –) |
| | Ireland (IRL) | 1000 | 4 (2, 2) | 1031 | 5 (0, 5) |
| | Kenya (KEN) | 972 | 4 (4, 0) | 984 | 4 (2, 2) |
| | New Zealand (NZL) | 1000 | 6 (3, 3) | 1031 | 5 (3, 2) |
| | Nigeria (NGA) | 1000 | 5 (3, 2) | 1031 | 6 (2, 4) |
| | Philippines (PHI) | 1000 | 4 (2, 2) | 1031 | 4 (1, 3) |
| | Scotland (SCO) | 1000 | 4 (2, 2) | 1031 | 5 (1, 4) |
| | South Africa (ZAF) | 1000 | 5 (3, 2) | 1031 | 6 (1, 5) |
| | USA-Southeast (USA-SE) | 1000 | 4 (2, 2) | 1031 | 5 (2, 3) |
| Kiswahili | Kenya (KEN) | 1825 | 7 (0, 7) | 2383 | 9 (7, 2) |
| | Tanzania (TZN) | 1803 | 7 (3, 4) | 2325 | 8 (4, 4) |
| Korean | South Korea-Seoul (KOR-C) | 371 | 2 (2, 0) | 1697 | 8 (5, 3) |
| | South Korea-south (KOR-SE) | 371 | 3 (2, 1) | 1697 | 7 (3, 4) |

Table 6: Data and annotator statistics for SD-QA.

**Korean** is the most populous member of the Koreanic language family (the other two being Jeju and Yukchin) spoken throughout the Korean peninsula. Korean is relatively homogeneous and the dialects from different areas can be mutually intelligible to a great extent. Nevertheless, the dialects of Korean exhibit considerable variety in phonology, morphology, and vocabulary. Most scholars agree on a division into six broad varieties (Sohn, 2001). In this work we collect data for the Central variety spoken in Seoul (considered "standard" Korean in the modern times) and for the Southeastern variety spoken in the Gyeongsang province. Unlike the Central dialect, the Southeastern variety has preserved the tonal distinctions of Middle Korean (essentially, a distinction between high and low pitch).[14] In the Southeastern dialects, standard Korean [e] and [ɛ] have merged as [e], and standard u [ɨ] and e [ə] have merged as e [ə]. We direct the reader to Yeon (2012) for a general survey of modern Korean varieties.

## C  Test Set Results

ASR results in Table 9, passage retrieval results in Table 10, and minimal answer results in Table 11.

## D  ASR Systems Unfairness

We use the framework described in Section §3 to quantify the unfairness of the models. We will limit our discussion here on English, where we have more than one model to compare, and provide extensive results in Appendix G. The two models we compare for English are the "general" one (obtained by using the en-US Google model) and a hypothetical "localized" one which assumes a pipeline involving dialect recognition (or a user-defined preference) that then selects the ASR model corresponding to the user's dialect.

---

[14]The non-tonal varieties exhibit vowel length differences as a trace of the tonal distinctions of Middle Korean.

| Language | Dialect | Population (million) |
|---|---|---|
| Arabic | Algeria (DZA) | 29 |
| | Bahrain (BHR) | 1 |
| | Egypt (EGY) | 62.3 |
| | Jordan (JOR) | 3.59 |
| | Morocco (MAR) | 24.8 |
| | Saudi Arabia (SAU) | 14.1 |
| | Tunisia (TUN) | 10.8 |
| Bengali | Bangladesh (BGD) | 163 |
| | India (IND) | 80 |
| English | Australia (AUS) | 25.36 |
| | India-South (IND-S) | 39.41 |
| | India-North (IND-N) | 154.69 |
| | Ireland (IRL) | 4.904 |
| | Kenya (KEN) | 52.6 |
| | New Zealand (NZL) | 4.917 |
| | Nigeria (NGA) | 178 |
| | Philippines (PHI) | 64 |
| | Scotland (SCO) | 5.46 |
| | South Africa (ZAF) | 58.56 |
| | USA-Southeast (USA-SE) | 125.6 |
| Kiswahili | Kenya (KEN) | 52.6 |
| | Tanzania (TZN) | 58 |
| Korean | South Korea-Seoul (KOR-C) | 47 |
| | South Korea-south (KOR-SE) | 2.8 |

Table 7: Population statistics for SD-QA dialectal regions.

| Dhaka | Kolkata | English |
|---|---|---|
| দোয়া *(doa)* | প্রার্থনা *(prarthona)* | pray |
| পানি *(pani)* | জল *(jol)* | water |
| দাওয়াত *(daowat)* | নিমন্ত্রণ *(nimontron)* | invitation |

Table 8: Example of lexical differences between the Dhaka and Kolkata Bengali varieties.

Our results, summarily displayed in Table 12, are that the "general" English model gets an unfairness score of 0.00191, while the "localized" one obtains a lower unfairness score of 0.00086. This means that the "localized" ASR model is not only slightly better in terms of average transcription quality, but it leads to a more fair distribution of its benefits among the underlying users. Further analysis of the within-group unfairness scores of each region for ASR shows that the highest performing region (New Zealand) also has the lowest unfairness score for the "localized" model.

This means that the "localized" ASR model is not only slightly better in terms of average transcription quality, but also leads to a more fair distribution of its benefits among the underlying users. Further analysis of the within-group and across-group unfairness scores of each region shows that both models achieve their lowest within-group unfairness scores for the US dialect (around 0.0014); this means that both models are more equitable for US English speakers than for speakers of other dialects. In contrast, the models have twice as high unfairness scores for South Indian English speakers (with unfairness scores around 0.0030). This means that not all South Indian English speakers receive consistent benefits (high quality transcriptions) by the systems.

| Model | English Variety | | | | | | | | | | | $\text{avg}_{\mathcal{L}}$ | $\text{avg}_{\text{pop}}$ |
| --- | --- | --- | --- | --- | --- | --- | --- | --- | --- | --- | --- | --- | --- |
| | AUS | IND-S | IND-N | IRE | NZL | NGA | PHI | SCO | ZAF | KEN | USA-SE | | |
| en-US | 10.43 | 11.65 | 10.13 | 9.25 | 8.79 | 15.35 | 8.32 | 11.35 | 8.57 | 12.81 | 5.25 | 10.17 | 10.57 |
| en-VAR | 8.54 | 16.48 | 13.29 | 8.91 | 7.27 | 11.61 | 8.95 | 8.14 | 7.56 | 10.24 | | 9.66 | 10.27 |

| Model | Bengali Variety | | | |
| --- | --- | --- | --- | --- |
| | BGD | IND | $\text{avg}_{\mathcal{L}}$ | $\text{avg}_{\text{pop}}$ |
| bn-BD | 30.87 | 32.30 | 31.59 | 31.34 |
| bn-IN | 30.83 | 32.30 | 31.56 | 31.31 |

| Model | Korean Variety | | | |
| --- | --- | --- | --- | --- |
| | KOR-C | KOR-SE | $\text{avg}_{\mathcal{L}}$ | $\text{avg}_{\text{pop}}$ |
| ko-KO | 26.13 | 26.45 | 26.29 | 26.15 |

| Model | Kiswahili Variety | | | |
| --- | --- | --- | --- | --- |
| | KEN | TZA | $\text{avg}_{\mathcal{L}}$ | $\text{avg}_{\text{pop}}$ |
| sw-KE | 36.63 | 44.74 | 40.69 | 40.88 |
| sw-TZ | 36.63 | 44.73 | 40.68 | 40.88 |

| Model | Arabic Variety | | | | | | | $\text{avg}_{\mathcal{L}}$ | $\text{avg}_{\text{pop}}$ |
| --- | --- | --- | --- | --- | --- | --- | --- | --- | --- |
| | DZA | BHR | EGY | JOR | MAR | SAU | TUN | | |
| ar-VARIETY | 37.84 | 37.93 | 40.40 | 38.55 | 39.09 | 38.24 | 38.70 | 38.68 | 39.27 |

Table 9: Test WER on each of the language varieties, using different localized speech recognition models.

Taking a look at the unfairness scores for the other languages, we observe that (a) all other models not only have worse ASR quality but are also more unfair over their respective populations. Bengali receives an unfairness score of 0.0355 while Arabic and Korean receive scores around 0.08. Kiswahili, however, beyond being by far the worst in terms of WER, is also the most unfair system with double the unfairness score (0.1626) of the second most unfair language system. This is unsurprising, considering how wide the performance (WER) gap is between Kenyan and Tanzanian Kiswahili.

**Sensitive Feature Analysis** The metadata associated with each annotator in SD-QA allow us to perform analyses across sensitive features like gender or age. We provide a breakdown of WER across these two features for all varieties and ASR models in Tables 14 (age) and 13 (gender) in Appendix E. The tables also provide information on the support for each average score to allow for better interpretation of the results. We leave a more sophisticated analysis incorporating statistical significance tests for future work.

Studying the effect of the speaker's gender, we do not find large differences between the average WER for most varieties, but nevertheless we can make some interesting observations. First, we note that Indian English behaves differently depending on whether the speakers are from south or north India. In particular, female speakers from the south receive slightly higher (worse) WER than their male counterparts (c.f. WER of 14.4 and 12.9). For speakers from the Indian north the situation is reversed, with female speakers receiving almost half of the WER of males (c.g. 8.5 and 15.8). For both Scottish and US English, the utterances of female speakers seem to be easier to transcribe than male ones, with a difference of about 3 WER points in both cases. While in Bangladesh Bengali we do not observe significant differences between the average WER for female and male speakers, in Indian Bengali the difference is more than 5 WER points, with male WER being better (lower).

## E ASR Results Breakdown

### E.1 By Gender

Results in Table 13.

### E.2 By Age

Results in Table 14.

## F ASR and QA Quality Correlations

Listed in Table 15.

## G Detailed Unfairness Results

**ASR Unfairness** We present a complete breakdown of the unfairness calculations for each language and dialect on the ASR task. For each dialect, we report two numbers. First, the within-group unfairness score, which can be interpreted as answering the question "how fair/consistent is the ASR model for the speakers of this dialect?". Second, we report the between-group unfairness score, which provides information on the average benefit a subgroup/dialect receives relative to all other dialects; a negative value means that this dialect is treated unfairly with respect to the rest.

| Transc. | AUS | IND-S | IND-N | IRE | NZL | NGA | PHI | ZAF | KEN | SCO | USA-SE | $\text{avg}_\mathcal{L}$ | $\text{avg}_{\text{pop}}$ |
|---|---|---|---|---|---|---|---|---|---|---|---|---|---|
| | | | | | **English Variety** | | | | | | | | |
| Gold | | | | | | 58.7 | | | | | | | |
| en-US | 56.4 | 54.1 | 56.1 | 56 | 55.1 | 54.1 | 57.2 | 56.3 | 54.6 | 57 | 57.2 | 55.8 | 55.7 |
| en-VAR | 57.1 | 52 | 53.9 | 57.1 | 57.2 | 53.7 | 54.5 | 56.5 | 55.9 | 56.8 | | 55.6 | 54.9 |

| Transc. | BGD | IND | $\text{avg}_\mathcal{L}$ | $\text{avg}_{\text{pop}}$ |
|---|---|---|---|---|
| | **Bengali Variety** | | | |
| Gold | | 61.9 | | |
| bn-XX | 60.3 | 59.7 | 60 | 60.1 |

| Transc. | KOR-C | KOR-SE | $\text{avg}_\mathcal{L}$ | $\text{avg}_{\text{pop}}$ |
|---|---|---|---|---|
| | **Korean Variety** | | | |
| Gold | | 58.1 | | |
| ko-KO | 55.7 | 54.9 | 55.3 | 55.7 |

| Transc. | KEN | TZA | $\text{avg}_\mathcal{L}$ | $\text{avg}_{\text{pop}}$ |
|---|---|---|---|---|
| | **Kiswahili Variety** | | | |
| Gold | | 62.7 | | |
| sw-XX | 45.9 | 42.1 | 44 | 43.9 |

| Transc. | DZA | BHR | EGY | JOR | MAR | SAU | TUN | $\text{avg}_\mathcal{L}$ | $\text{avg}_{\text{pop}}$ |
|---|---|---|---|---|---|---|---|---|---|
| | **Arabic Variety** | | | | | | | | |
| Gold | | | | 79.9 | | | | | |
| ar-VAR | 77.9 | 77 | 76.3 | 77.7 | 77.5 | 76.6 | 77.3 | 77.2 | 76.9 |

Table 10: Baseline passage selection results ($F_1$-SCORE, higher is better) on the SD-QA test set.

Table 16 presents all results.

**QA Unfairness** Detailed unfairness results for both QA tasks are listed in Tables 17 and 18.

## H Further Related Work

**Benchmark Datasets** There exist a number of benchmark question answering datasets. SQuAD (Rajpurkar et al., 2016) provides a passage and a question which has an answer placed in the passage. In SQuAD 2.0 (Rajpurkar et al., 2018) additional unanswerable questions were introduced. However, in SQuAD the annotators first read the passage and create the questions based on it. A more realistic setting is to instead focus on "random" questions which might be asked to search engines without reading any passage. The Natural Questions (NQ) dataset (Kwiatkowski et al., 2019) attempts to address this gap, consisting of anonymized Google queries. Other datasets focus on a trivia (TriviaQA (Joshi et al., 2017)) or conversational setting (CoQA (Reddy, 1989) and QuAC (Choi et al., 2018)). Generally, conversational and dialog based QA datasets introduce new challenges as the questions are in free form and highly contextual. Notably, all the aforementioned datasets are only in English.

**Multilingual QA** Beyond monolingual QA models, cross-lingual QA systems aim to leverage resources from one language to answer questions originally asked in a different language. Recently released, the TyDi-QA dataset, upon which we build, contains question-passage-answer pairs in 11 typologically diverse languages. In (Debnath et al., 2021), the authors perform an empirical analysis on TyDi-QA gold passage task, where we can see few-shot learning and translation based cross-lingual transfer is an effective diverse dataset development approach with fixed annotation budget. XOR-QA (Asai et al., 2020) explores the direction of open domain QA systems by introducing 3 cross lingual tasks. Asai et al. (2020) also build their cross lingual dataset on top of TyDi-QA questions, where questions originally unanswerable are associated with useful resources like English translations, related English wikipedia articles and any answers found are then translated to the original question language. Other notable benchmark cross-lingual QA datasets include MLQA (Lewis et al., 2019), MKQA (Longpre et al., 2020) and XQuAD (Artetxe et al., 2019). MLQA (Lewis et al., 2019) also explores the questions being translations of the English ones from the SQuAd dataset. We opted for recording questions from the TyDi-QA dataset, in order to work with realistic questions for each language and avoid the effect of translationese.[15]

| Transc. | AUS | IND-S | IND-N | IRE | NZL | NGA | PHI | ZAF | KEN | SCO | USA-SE | $\text{avg}_\mathcal{L}$ | $\text{avg}_{\text{pop}}$ |
|---|---|---|---|---|---|---|---|---|---|---|---|---|---|
| | | | | | **English Variety** | | | | | | | | |
| Gold | | | | | | 37 | | | | | | | |
| en-US | 33.5 | 31.8 | 33.7 | 33.6 | 33.1 | 31.7 | 35.7 | 35 | 33.4 | 34 | 35.7 | 33.7 | 33.6 |
| en-VAR | 34.5 | 30.8 | 32 | 34.0 | 35.3 | 31.7 | 33.2 | 34 | 33 | 33.9 | | 33.5 | 33 |

| | **Bengali Variety** | | | | | **Korean Variety** | | | |
|---|---|---|---|---|---|---|---|---|---|
| Transc. | BGD | IND | $\text{avg}_\mathcal{L}$ | $\text{avg}_{\text{pop}}$ | Transc. | KOR-C | KOR-SE | $\text{avg}_\mathcal{L}$ | $\text{avg}_{\text{pop}}$ |
| Gold | 47.9 | | | | Gold | 39.6 | | | |
| bn-XX | 47.2 | 45.8 | 46.5 | 46.7 | kor-STD | 38.0 | 38.8 | 38.4 | 38.1 |

| | **Kiswahili Variety** | | | |
|---|---|---|---|---|
| Transc. | KEN | TZA | $\text{avg}_\mathcal{L}$ | $\text{avg}_{\text{pop}}$ |
| Gold | 50 | | | |
| sw-XX | 32.9 | 29.7 | 31.3 | 31.2 |

| | **Arabic Variety** | | | | | | | |
|---|---|---|---|---|---|---|---|---|
| Transc. | DZA | BHR | EGY | JOR | MAR | SAU | TUN | $\text{avg}_\mathcal{L}$ | $\text{avg}_{\text{pop}}$ |
| Gold | | | | 66.4 | | | | | |
| ara-VAR | 62.3 | 62.1 | 61.1 | 62.3 | 62.3 | 61.7 | 62.4 | 62.0 | 61.7 |

Table 11: Baseline results on the minimal answer task ($F_1$-SCORE, higher is better) on the SD-QA test set.

| ASR Model | Unfairness score ↓ | Avg. WER ↓ |
|---|---|---|
| en-US | 0.02456 | 12.17 |
| en-VAR | 0.02282 | 11.35 |
| ar-XX | 0.07701 | 36.36 |
| bn-XX | 0.03540 | 27.6 |
| ko-KO | 0.08674 | 26.29 |
| sw-XX | 0.11152 | 43.49 |

Table 12: A hypothetical English ASR model using localized ASR models is not only better in terms of average quality (WER) but also slightly more equitable than the "general" en-US model. The ASR systems for other languages are generally more unfair than the English ones.

cross lingual alignment among 7 language instead of training on monolingual large dataset, providing translations of the original English questions. (Bornea et al., 2020) extends MLQA by introducing translation based data augmentation and adversarial training. They further improved the performance of TyDi-QAmodel in crosslingual setting by introducing additional loss function to provide similar prediction for translated questions. MKQA (Longpre et al., 2020) provides the most diverse multilingual QA dataset comprised of 26 languages, with the data being translations of English NQ questions. Last, XQuAD (Artetxe et al., 2019) is another translated benchmark, with

---

[15]See discussion on "Why not translate?" by Clark et al. (2020).

| Model | Gender | \multicolumn{11}{c}{English Variety} |
| | | AUS | IND-S | IND-N | IRE | NZL | NGA | PHI | ZAF | KEN | SC0 | USA-SE |
|---|---|---|---|---|---|---|---|---|---|---|---|---|
| en-US | Female | 10.92 (446) | 14.36 (867) | 8.55 (247) | 9.1 (517) | 10.28 (372) | 15.16 (455) | 12.4 (535) | 12.3 (512) | – | 9.46 (434) | 6.92 (457) |
| | Male | 10.96 (526) | 12.93 (105) | 15.82 (725) | 10.34 (455) | 8.9 (600) | 14.15 (517) | 12.89 (437) | 10.7 (460) | 13.61 (912) | 12.58 (538) | 10.83 (515) |
| en-VAR | Female | 9.65 (446) | 17.03 (867) | 11.66 (247) | 8.88 (517) | 8.86 (372) | 11.49 (455) | 12.59 (535) | 9.23 (512) | – | 7.52 (434) | |
| | Male | 11.06 (526) | 18.6 (105) | 19.24 (725) | 10.73 (455) | 8.58 (600) | 10.72 (517) | 11.93 (437) | 8.18 (460) | 11.13 (912) | 10.69 (538) | |

| Model | Gender | Bengali Variety | |
| | | BGD | IND |
|---|---|---|---|
| bn-BD | Female | 28.46 (180) | 31.58 (63) |
| | Male | 28.38 (247) | 25.86 (364) |
| bn-IN | Female | 28.46 (180) | 31.58 (63) |
| | Male | 28.49 (247) | 25.86 (364) |

| Model | Gender | Korean Variety | |
| | | KOR-C | KOR-SE |
|---|---|---|---|
| kor-STD | Female | – | 20.73 (33) |
| | Male | 27.81 (370) | 24.92 (337) |

| Model | Gender | Kiswahili Variety | |
| | | KEN | TZA |
|---|---|---|---|
| sw-KE | Female | 32.0 (1743) | 40.27 (1025) |
| | Male | 27.3 (48) | 43.78 (765) |
| sw-TZ | Female | 32.02 (1743) | 40.27 (1025) |
| | Male | 27.3 (48) | 43.74 (765) |

| Model | Gender | \multicolumn{7}{c}{Arabic Variety} |
| | | DZA | BHR | EGY | JOR | MAR | SAU | TUN |
|---|---|---|---|---|---|---|---|---|
| ar-VAR | Female | 39.68 (168) | 35.01 (270) | 38.64 (168) | 33.81 (270) | 34.67 (672) | 36.48 (540) | 34.2 (270) |
| | Male | 37.07 (540) | 35.9 (438) | 36.64 (540) | 36.22 (438) | 41.7 (36) | 33.37 (168) | 35.28 (438) |

Table 13: Development WER (example count grouped by speaker gender) for each of the language varieties, using dialect specific and general speech recognition.

| Transc. | Age | \multicolumn{11}{c}{English Variety} |
| | | AUS | IND-S | IND-N | IRE | NZL | NGA | PHI | ZAF | KEN | SCO | USA-SE |
|---|---|---|---|---|---|---|---|---|---|---|---|---|
| en-US | 18-30 | 9.53 (186) | 16.18 (452) | 18.18 (540) | 7.46 (190) | 9.92 (517) | 13.49 (592) | 12.13 (702) | 11.86 (80) | 13.68 (972) | – | 7.96 (455) |
| | 31-45 | 12.35 (529) | 12.54 (520) | 8.58 (432) | 10.26 (782) | 8.91 (440) | 16.3 (380) | – | 11.04 (892) | – | 11.18 (972) | 9.88 (517) |
| | 46-59 | 8.99 (257) | – | – | – | 7.92 (15) | – | 13.95 (270) | – | – | – | – |
| en-VAR | 18-30 | 10.04 (186) | 19.12 (452) | 19.71 (540) | 7.74 (190) | 8.49 (517) | 10.82 (592) | 12.19 (702) | 11.86 | 11.04 (972) | – | |
| | 31-45 | 11.76 (529) | 15.57 (520) | 14.21 (432) | 10.28 (782) | 8.91 (440) | 11.45 (380) | – | 8.43 (892) | – | 9.27 (972) | |
| | 46-59 | 7.82 (257) | – | – | – | 8.91 (15) | – | 12.53 (270) | – | – | – | |

| Transc. | Age | Bengali Variety | |
| | | BGD | IND |
|---|---|---|---|
| bn-BD | 18-30 | 28.71 (397) | 25.7 (270) |
| | 31-45 | 24.78 (30) | 27.92 (151) |
| | 46-59 | – | 39.53 (6) |
| bn-IN | 18-30 | 28.78 (397) | 25.7 (270) |
| | 31-45 | 24.78 (30) | 27.92 (151) |
| | 46-59 | – | 39.53 (6) |

| Transc. | Age | Korean Variety | |
| | | KOR-C | KOR-SE |
|---|---|---|---|
| kor-STD | 18-30 | – | 20.73 (33) |
| | 31-45 | 29.11 (269) | 25.04 (269) |
| | 46-59 | 24.15 (101) | 24.44 (68) |

| Trans. | Age | Kiswahili Variety | |
| | | KEN | TZA |
|---|---|---|---|
| sw-KE | 18-30 | 47.7 (300) | 45.69 (1248) |
| | 31-45 | 27.02 (714) | 32.06 (542) |
| | 46-59 | 30.23 (777) | – |
| sw-TZ | 18-30 | 47.7 (300) | 45.69 (1248) |
| | 31-45 | 27.06 (714) | 32.0 (542) |
| | 46-59 | 30.23 (777) | – |

| Model | Age | \multicolumn{7}{c}{Arabic Variety} |
| | | DZA | BHR | EGY | JOR | MAR | SAU | TUN |
|---|---|---|---|---|---|---|---|---|
| ara-VAR | 18-30 | 39.68 (168) | 35.01 (270) | – | 35.32 (708) | 35.06 (708) | 36.48 (540) | 32.91 (438) |
| | 31-45 | 37.07 (540) | 32.34 (168) | 36.33 (270) | – | – | 33.37 (168) | 37.81 (270) |
| | 46-59 | – | 37.92 (270) | 37.63 (438) | – | – | – | – |

Table 14: Development WER (example count by speaker age-group) for each of the language varieties, using dialect specific and general speech recognition.

| Language | Passage Sel. | Minimal Answer |
|---|---|---|
| ara | -0.29 | -0.25 |
| ben | -0.94 | -0.94 |
| eng | -0.92 | -0.81 |
| swa | -0.74 | -0.74 |
| kor | -1.00 | -1.00 |

Table 15: Speech transcription quality correlates with downstream QA accuracy (Spearman's rank correlation coefficient between WER and $F_1$-SCORE). Since for WER lower is better, but for QA $F_1$-SCORE higher is better, the correlations are negative.

| Model | Total Unfairness | Component | AUS | IND-S | IND-N | IRE | English Variety NZL | NGA | PHI | ZAF | KEN | SC0 | USA-SE |
|---|---|---|---|---|---|---|---|---|---|---|---|---|---|
| en-US | 0.02456 | within group between group | 0.00186 | 0.00293 | 0.00277 | 0.00172 | 0.00165 | 0.00282 0.00031 | 0.00238 | 0.00224 | 0.00243 | 0.00197 | 0.00148 |
| en-VAR | 0.02282 | within group between group | 0.00193 | 0.00341 | 0.00322 | 0.00176 | 0.00167 | 0.00240 0.00067 | 0.00229 | 0.00176 | 0.00198 | 0.00171 | - |

| Model | Total Unfairness | Component | Bengali Variety BGD | IND |
|---|---|---|---|---|
| bn-BD | 0.03540 | within group between group | 0.01873 0.00006 | 0.01662 |
| bn-IN | 0.03550 | within group between group | 0.01882 0.00006 | 0.01663 |

| Model | Total Unfairness | Component | Korean Variety KOR-C | KOR-SE |
|---|---|---|---|---|
| kor-STD | 0.08674 | within group between group | 0.04272 0.00007 | 0.04396 |

| Model | Total Unfairness | Component | Kiswahili Variety KEN | TZA |
|---|---|---|---|---|
| sw-KE | 0.11152 | within group between group | 0.04874 0.00343 | 0.05935 |
| sw-TZ | 0.11158 | within group between group | 0.04876 0.00340 | 0.05942 |

| Model | Total Unfairness | Component | DZA | BHR | Arabic Variety EGY | JOR | MAR | SAU | TUN |
|---|---|---|---|---|---|---|---|---|---|
| ar-VAR | 0.07701 | within group between group | 0.01200 | 0.01090 | 0.01131 0.00013 | 0.01091 | 0.01065 | 0.01049 | 0.01060 |

Table 16: Development unfairness and components for ASR, computed using 1-WER as benefit.

| Model | Total Unfairness | Component | AUS | IND-S | IND-N | IRE | English Variety NZL | NGA | PHI | ZAF | KEN | SC0 | USA-SE |
|---|---|---|---|---|---|---|---|---|---|---|---|---|---|
| en-US | 0.08151 | within group between group | 0.00751 | 0.00705 | 0.00737 | 0.00755 | 0.00769 | 0.00775 0.00016 | 0.00730 | 0.00719 | 0.00694 | 0.00748 | 0.00752 |
| en-VAR | 0.08226 | within group between group | 0.00822 | 0.00759 | 0.00765 | 0.00834 | 0.00902 | 0.00835 0.00013 | 0.00824 | 0.00801 | 0.00839 | 0.00833 | - |

| Model | Total Unfairness | Component | Bengali Variety BGD | IND |
|---|---|---|---|---|
| bn-BD | 0.04322 | within group between group | 0.02057 0.00070 | 0.02195 |
| bn-IN | 0.04164 | within group between group | 0.01998 0.00028 | 0.02138 |

| Model | Total Unfairness | Component | Korean Variety KOR-C | KOR-SE |
|---|---|---|---|---|
| kor-STD | 0.05528 | within group between group | 0.02773 0.00001 | 0.02754 |

| Model | Total Unfairness | Component | Kiswahili Variety KEN | TZA |
|---|---|---|---|---|
| sw-KE | 0.08903 | within group between group | 0.04494 0.00000 | 0.04409 |
| sw-TZ | 0.08908 | within group between group | 0.04507 0.00000 | 0.04401 |

| Model | Total Unfairness | Component | DZA | BHR | Arabic Variety EGY | JOR | MAR | SAU | TUN |
|---|---|---|---|---|---|---|---|---|---|
| ar-VAR | 0.07606 | within group between group | 0.01058 | 0.01099 | 0.01066 0.00012 | 0.01086 | 0.01128 | 0.01069 | 0.01089 |

Table 17: Development unfairness and components for passage selection QA task, computed using F1-score as benefit.

| Model | Total Unfairness | Component | English Variety | | | | | | | | | | |
|---|---|---|---|---|---|---|---|---|---|---|---|---|---|
| | | | AUS | IND-S | IND-N | IRE | NZL | NGA | PHI | ZAF | KEN | SC0 | USA-SE |
| en-US | 0.07790 | within group | 0.00730 | 0.00664 | 0.00723 | 0.00719 | 0.00734 | 0.00678 | 0.00664 | 0.00702 | 0.00636 | 0.00724 | 0.00749 |
| | | between group | | | | | | 0.00067 | | | | | |
| en-VAR | 0.07832 | within group | 0.00785 | 0.00685 | 0.00680 | 0.00789 | 0.00836 | 0.00831 | 0.00818 | 0.00795 | 0.00749 | 0.00812 | - |
| | | between group | | | | | | 0.00053 | | | | | |

| Model | Total Unfairness | Component | Bengali Variety | |
|---|---|---|---|---|
| | | | BGD | IND |
| bn-BD | 0.04686 | within group | 0.02382 | 0.02299 |
| | | between group | 0.00004 | |
| bn-IN | 0.04501 | within group | 0.02276 | 0.02222 |
| | | between group | 0.00003 | |

| Model | Total Unfairness | Component | Korean Variety | |
|---|---|---|---|---|
| | | | KOR-C | KOR-SE |
| kor-STD | 0.03980 | within group | 0.02049 | 0.01928 |
| | | between group | 0.00002 | |

| Model | Total Unfairness | Component | Kiswahili Variety | |
|---|---|---|---|---|
| | | | KEN | TZA |
| sw-KE | 0.09145 | within group | 0.04918 | 0.04209 |
| | | between group | 0.00018 | |
| sw-TZ | 0.09153 | within group | 0.04933 | 0.04200 |
| | | between group | 0.00021 | |

| Model | Total Unfairness | Component | Arabic Variety | | | | | | |
|---|---|---|---|---|---|---|---|---|---|
| | | | DZA | BHR | EGY | JOR | MAR | SAU | TUN |
| ar-VAR | 0.07602 | within group | 0.01049 | 0.01102 | 0.01075 | 0.01072 | 0.01134 | 0.01072 | 0.01092 |
| | | between group | | | | 0.00007 | | | |

Table 18: Development unfairness and components for minimal answer QA task, computed using F1-score as benefit.